# SURFACE NORMAL RECONSTRUCTION USING POLARIZATION-UNET

F. S. Mortazavi [1]*, S. Dajkhosh[2], M. SaadatSeresht[1]

[1] School of Surveying and Geospatial Engineering, College of Engineering, University of Tehran, Tehran, Iran – (faezeh.mortazavi, msaadat)@ut.ac.ir
[2] School of Mathematics, Statistics and Computer Science, College of Science, University of Tehran, Tehran, Iran – parsadaj@ut.ac.ir

**Commission IV, WG IV/3**

**ABSTRACT:**

Today, three-dimensional reconstruction of objects has many applications in various fields, and therefore, choosing a suitable method for high resolution three-dimensional reconstruction is an important issue and displaying high-level details in three-dimensional models is a serious challenge in this field. Until now, active methods have been used for high-resolution three-dimensional reconstruction. But the problem of active three-dimensional reconstruction methods is that they require a light source close to the object. Shape from polarization (SfP) is one of the best solutions for high-resolution three-dimensional reconstruction of objects, which is a passive method and does not have the drawbacks of active methods. The changes in polarization of the reflected light from an object can be analyzed by using a polarization camera or locating polarizing filter in front of the digital camera and rotating the filter. Using this information, the surface normal can be reconstructed with high accuracy, which will lead to local reconstruction of the surface details. In this paper, an end-to-end deep learning approach has been presented to produce the surface normal of objects. In this method a benchmark dataset has been used to train the neural network and evaluate the results. The results have been evaluated quantitatively and qualitatively by other methods and under different lighting conditions. The MAE value (Mean-Angular-Error) has been used for results evaluation. The evaluations showed that the proposed method could accurately reconstruct the surface normal of objects with the lowest MAE value which is equal to 18.06 degree on the whole dataset, in comparison to previous physics-based methods which are between 41.44 and 49.03 degree.

## 1. INTRODUCTION

Today, three-dimensional reconstruction of objects has many applications in various fields, and therefore, choosing a suitable method for high resolution three-dimensional reconstruction is an important issue. There are many ways to classify three-dimensional reconstruction methods. Common ways to classify methods include classification based on whether the method is active or passive. Active three-dimensional reconstruction methods are methods that reconstruct a three-dimensional scene using approximate numerical methods according to the depth map. These methods use light radiation or irradiation and the object is reconstructed with the help of the reflection of these rays. Passive three-dimensional reconstruction methods do not interfere with the object being reconstructed and only use sensors that are sensitive to visible light to measure the radiation reflected or emitted from the surface of the object to infer its three-dimensional structure through image analysis. The input of these methods is a series of digital images and the output of this method is a three-dimensional model. Passive methods are used in a wide range of situations compared to active methods (Moons et al., 2010).

Combined solutions have been proposed to achieve high resolution three-dimensional reconstruction, such as combining an initial three-dimensional model with details obtained from active methods such as shape from structured light (Nehab et al., 2005), photometric stereo (Esteban et al., 2008; Park et al., 2013; Zhou et al., 2013; Haque et al., 2014), or shape from shading (Wu et al., 2011; Oxholm and Nishino, 2014; Langguth et al., 2016). Recently, appropriate solutions have been proposed for high resolution three-dimensional reconstruction of objects without using active methods. Shape from polarization method is one of these solutions which is a passive method and does not have the drawbacks of active methods. This method is based on the concept that the shape of an object causes small changes in the polarization of the reflected light. The electric fields that build up light waves are randomly placed in any direction, but in light polarization, the electric field is only sent or received in a certain direction. These changes can be analyzed by locating a polarizing filter in front of the digital camera and rotating the filter (Wolff, 1997; Atkinson and Ernst, 2018) or using a polarization camera (Polarization_camera, 2020; Yang et al., 2018). Through these polarization images, the surface normal can be reconstructed with high accuracy, which will lead to local reconstruction of the surface details.

The shape from polarization method has made considerable progress in the field of computer vision. One of the most common solutions is to combine an initial three-dimensional model with the details obtained from shape from polarization method. In this integrated solution, a surface is first reconstructed using polarization information and Fresnel theory, which reveals the surface details well. By combining this surface with the surfaces obtained from conventional photogrammetric methods, depth maps can be significantly enhanced using information obtained from the polarization of the emitted light. In this paper, a method is presented that reconstructs surface normal of objects using a convolutional neural network and only using polarized images, without using any additional information.

---

\* Corresponding author

## 2. RELATED WORKS

Recently, many methods have been proposed for three-dimensional reconstruction based on light polarization. Some of these researches have only used the physics of polarization without using other methods for three-dimensional reconstruction of objects. In general, in the polarization method, information such as phase angle, degree of polarization and zenith angle can be obtained from polarization images, and finally, using this information, the surface normal of the object can be estimated. The phase angle obtained from this method is ambiguous, so some researchers have proposed methods to solve this phase ambiguity (Miyazaki et al., 2003; Atkinson and Hancock, 2006). This method is a suitable solution for the reconstruction of transparent objects. For example, Miyazaki et al. (2002) has presented a method that shows the orientation of the surface in transparent objects by analyzing the degree of polarization in the surface reflection and its propagation in visible and infrared wavelengths. This method can also perform well in the reconstruction of specular metallic objects (Morel et al., 2005).

Overall, the polarization method cannot be used alone and still has ambiguities and drawbacks. For this reason, one of the appropriate solutions is to combine SfP method with conventional photogrammetric methods for three-dimensional reconstruction. In this case, the weaknesses of the polarization method can be covered with photogrammetric methods. Therefore, the researchers proposed combined methods, for example, one of these methods uses a combination of polarization, stereo, and shape from shading methods (Zhu and Smith, 2019). Tozza et al. (2017) presented a differential method combining polarization and shading to reconstruct depth maps. In another work, Atkinson and Hancock (2007b) proposed a combination of polarization, shadow, and stereo methods. Methods for resolving ambiguity and reconstructing the surface normal using a combination of polarization and photometric stereo are also presented (Atkinson and Hancock, 2007a; Atkinson 2017). Miyazaki et al. (2012 and 2016) used multi view space carving to achieve surface normal for improving results in black specular objects. Ngo Thanh et al. (2015) used two constraints of shading and polarization to solving phase ambiguity as well as estimating the refractive index, normal surface and light direction. Mahmoud et al. (2012) and Smith et al. (2016) have proposed direct methods that use polarization and shading methods that assumes constant orthographic projection and albedo. Next, Smith et al. (2018) improved their previous work (Smith et al., 2016) using albedo estimation and illumination using polarization information. The polarization method can work well in feature correspondence matching, so it can be useful for matching of featureless areas in multiview stereo methods (Atkinson and Hancock, 2005; Cui et al., 2017). This method can also be combined with coarse depth maps obtained from Kinect and the accuracy can be significantly improved (Kadambi et al., 2017).

However, the polarization method is still immature and has some ambiguities. Therefore, one of the best solutions can be to use deep learning methods for three-dimensional reconstruction of objects using polarization images. The first research in this field has been conducted by Ba et al. (2020), that achieves a normal surface using CNN networks. The CNN networks have also been used to segment transparent objects (Kalra et al., 2020). Also, the deep learning network has been used to overcome the physics-based drawbacks of the polarization method (Deschaintre et al., 2021). In the latest research, for the first time, Lei et al., (2021) uses real-world dataset for reconstruction.

## 3. PROPOSED METHOD

### 3.1 Polarization Theory

When unpolarized light passes through a polarization filter, the light passes linearly only in the direction of the filter transmission axis, and the light is removed in other directions (Figure.1).

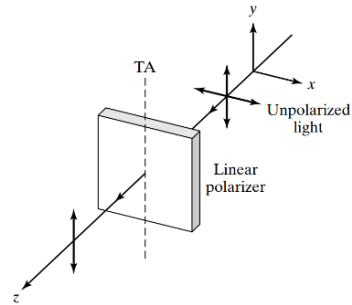

**Figure 1**. Operation of a linear polarizer (Pedrotti et al., 2017)

Polarized images can be taken at different angles using a polarizing filter located in the front of the digital camera and rotated. Using these polarization images, information such as azimuth and zenith angle can be obtained which leads to estimation of surface normal.

At least 3 polarized images are required to obtain polarization information. In this series of images, the intensity of each pixel changes sinusoidally between $I_{min}$ and $I_{max}$ by changing the polarization angle. A polarized image is taken at the angle $\varphi_{pol}$. The intensity of each pixel in each polarized image can be obtained from Equation 1:

$$I(\varphi_{POL}) = \frac{I_{max} + I_{min}}{2} + \frac{I_{max} - I_{min}}{2} \cos(2(\varphi_{POL} - \varphi)) \quad (1)$$

In this equation, the three parameters $I_{min}$, $I_{max}$ and $\varphi$ are unknown. The phase angle obtained from this relationship is ambiguous and this ambiguity varies depending on the type of pixel reflection that is diffuse or specular. If the pixel reflection is diffuse, this ambiguity consists of two values $\varphi$ and $\varphi + \pi$, and if its reflection is specular, it has the values $\varphi \pm \frac{\pi}{2}$. In fact, the azimuth angle, after solving the phase angle ambiguity, has one of the following values:

$$\phi = \varphi \pm \frac{\pi}{2} \quad \text{and} \quad \phi = \varphi \text{ or } \varphi + \pi \quad (2)$$

To obtain the zenith angle, the value of degree of polarization is required, which can be obtained from the minimum and maximum intensity:

$$\rho = \frac{I_{max} - I_{min}}{I_{max} + I_{min}} \quad (3)$$

Different relations are used to obtain the zenith angle depending on whether the pixel is specular or diffuse. If the pixel is diffuse, the value of the zenith angle is obtained from Equation 4, and if it is specular, it is obtained from Equation 5.

In these equations, $\eta$ represents the refractive index and $\theta$ is the zenith angle.

$$\rho = \frac{(\eta - \frac{1}{\eta})^2 \sin^2\theta}{2 + 2\eta^2 - (\eta + \frac{1}{\eta})^2 \sin^2\theta + 4\cos\theta\sqrt{\eta^2 - \sin^2\theta}} \quad (4)$$

$$\rho^{spec} = \frac{2\sin^2\theta \cos\theta\sqrt{\eta^2 - \sin^2\theta}}{\eta^2 - \sin^2\theta - \eta^2\sin^2\theta + 2\sin^4\theta} \quad (5)$$

Since phase disambiguation has always been one of the main challenges of this method, in this paper we use a convolutional neural network that will not face these problems.

### 3.2 Network architecture

U-Net is one of the famous architectures that was first used in 2015 for segmentation of biomedical images. The U-Net consists of two main parts: contracting path for feature extraction and expanding path for up-sampling and output generation (Ronneberger et al., 2015). Since in this work the outputs will be produced with the same size of the inputs, the U-Net is an appropriate choice. The well-known ResNet18 (He et al., 2015) is utilized as the backbone of this network to extract geometric and semantic features.

**3.2.1 U-Net:** In the contracting path, which is actually the backbone network, feature extraction is performed, so that during this path, the dimensions of the image and feature maps are gradually reduced, and instead, the number of these feature maps increases. At the end of this path, it is expected to extract geometric and semantic high-level features. Now these extracted features are entered into the expanding path so that the desired output is obtained from these obtained features. Hence, several layers of convolution and up-sampling are used in such a way that first a 2 x 2 up-convolution layer is used to double the size of the image and halve the number of feature maps. Since the features produced in this step (due to up-sampling) have little spatial information, the lower-level feature maps that exist in the contracting path are used. After each up-sampling, these features are concatenated with the features generated from the previous layer and are used as input to the next layer. After the up-sampling layer, two 3 x 3 convolution layers are applied to prepare the feature maps for entering the next up-sampling layer. This continues until the dimensions of the feature maps are equal to the dimensions of the expected output. In this step, the required output is obtained by using a 3 x 3 convolution and linear activation function.

**3.2.2 ResNet18:** ResNet became the talk of the town by achieving breathtaking results on ImageNet dataset (Deng et al., 2009) in 2015 while having a large number of layers.
Previously, vanishing and exploding gradient problem was a huge barrier to use a lot of layers (Glorot et al., 2010). ResNet solved this issue pretty well by introducing residual blocks. A residual block allows the network to pass the input feature alongside with the outputs to the next block and helps the network to learn an identity function easier when needed.

$$H(x) = x + F(x) \quad (6)$$

As shown in Equation 6, the input feature (x) is added to result of the convolution block F(x) to produce the output feature H(x) (He et al., 2015).

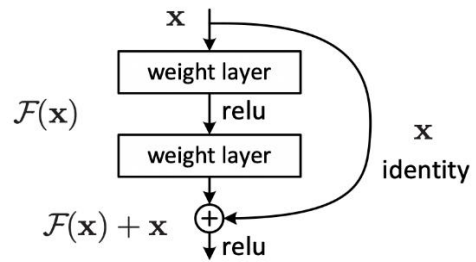

**Figure 2**. A residual block (He et al., 2015)

In this paper, we use ResNet18 as the backbone of the U-Net, which consists of 18 layers.
An overview of the proposed model has been shown in Figure 3. At each ResNet convolutional stage, the height and width of the input feature map are halved and the number of feature maps is doubled, except for the first and the second stages in which they both have 64 feature maps.
In the decoder part, which is on the right side of the figure, up-sampling layers double the size of the feature maps while it reduces the number of the feature maps to half. In the final part, a 3 x 3 convolution layer with linear activation is performed to produce the outputs.

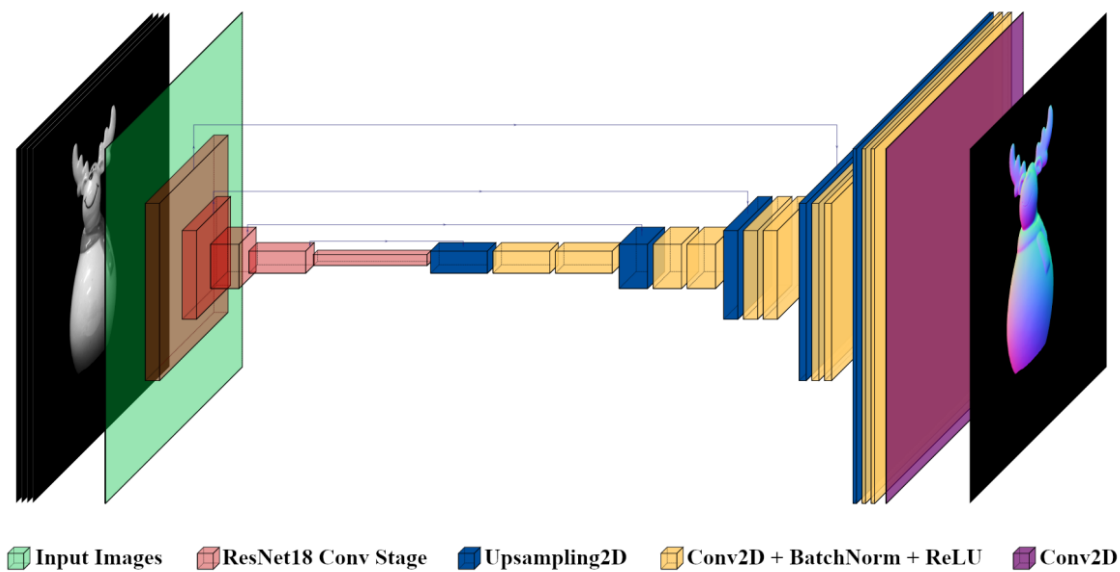

**Figure 3**. The overview of the network

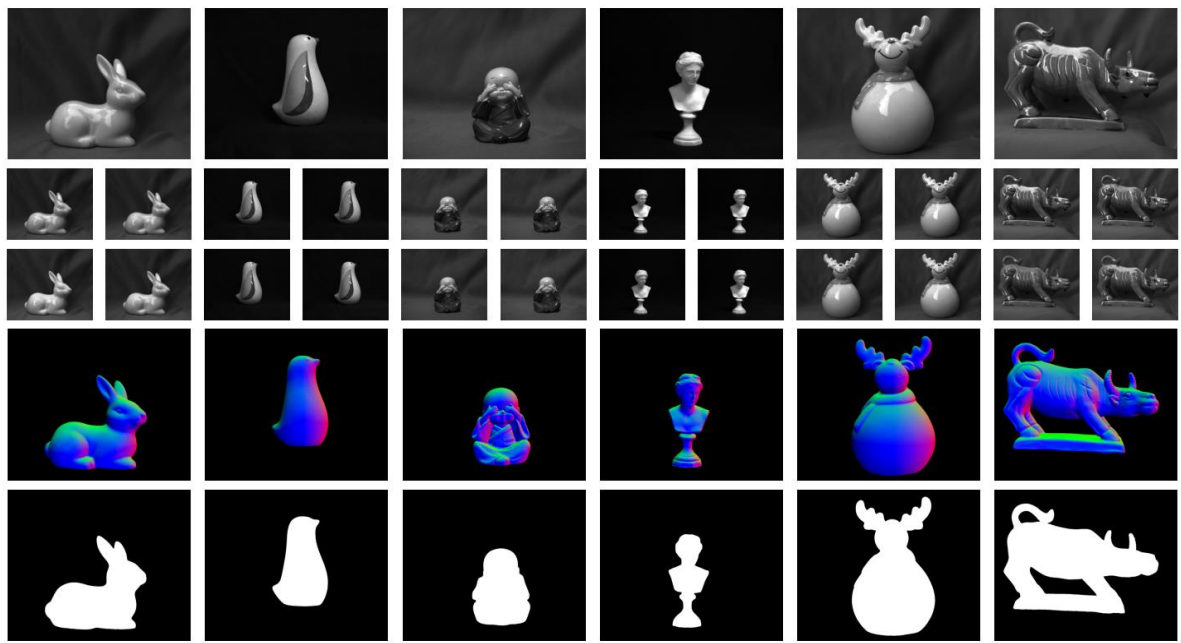

**Figure 4**. Six samples of dataset which include polarized images in 4 different degrees (0, 45, 90 and 135) in second row, ground truth in third row and binary mask in fourth row.

## 4. DATA AND RESULTS

### 4.1 Dataset

In this section, the details of the used dataset along with software and hardware specifications are explained. Deep Shape from Polarization dataset (Ba et al., 2020) is used for both training and testing. For each object at most 12 different training samples are created in three different lightning conditions (i.e., indoor, sunny outdoor, and cloudy outdoor) and from four different views (i.e., front, back, left, and right). Each sample contains three items:

1. **a polarized image** (1024x1224x4) in which each dimension represents the image in a specific polarization angle (0, 45, 90, and 135 degrees respectively)
2. **a surface normal image** (1024x1224x3) including x, y, and z components of the surface normal vector for each pixel
3. **a binary mask** (1024x1224x1) that separates the foreground pixels from the background.

Train and test split is performed in such a way that 25 objects (235 samples) in training set and 8 objects (27 samples) in test set. Then, 64x64 patches are extracted from 1024x1224 samples to get fed into the network. At the end, there are 27190 samples in the training set and 3254 samples in the test set. In the training phase, 20% of the training data are detached to be used as the validation data.

### 4.2 Training

Keras framework (Chollet et al., 2015) with NVIDIA GeForce RTX 2060 GPU with 6GB of VRAM is used for training the network. Adam optimizer (Kingma and Ba, 2014) with 0.0001 learning rate is chosen to find the optimum weights within 100 epochs, and training data are fed into the network in batch size 32. In order to having a better generalization of model, an L2 regularization with regularization factor of 1 has been used for all convolutional layers. The learning curves have been shown in Figure 5.

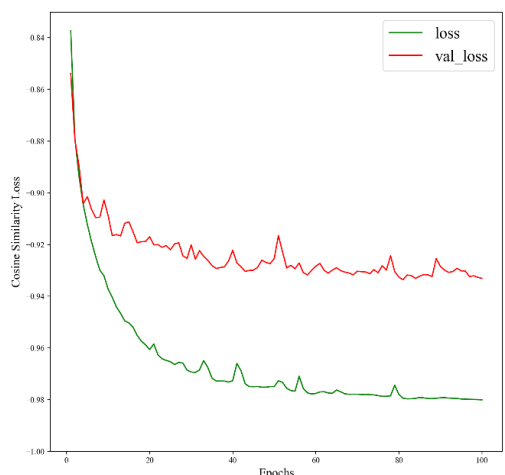

**Figure 5**. Cosine similarity loss curve in the training process for train and validation data

### 4.3 Assessments

In this paper, the results are evaluated both quantitatively and qualitatively.

**4.3.1 Qualitative assessment**: Figure 6 shows the surface normal of six objects (Father Christmas, Flamingo, Horse, Dragon, Box, and Vase) in three different lighting conditions (indoor, sunny outdoor, and cloudy outdoor). The MAE value of these objects has been written on the top left corner of surface normal images.

Quantitative and qualitative results in these images show that the method can reconstruct the surface normal of objects with the lowest MAE value. In addition, by assessing the MAE values in different lighting conditions, it is clear that this method was able to reconstruct the object in different conditions with the less variations.

**4.3.2 Quantitative assessment**: MAE (Mean Angular Error) metric has been used for quantitative evaluation in this paper. MAE metric is the most common way for measurement of surface normal reconstruction and to evaluate the difference between estimated surface normal and its ground true value. Table 1 shows the results of the proposed method along with other recently developed methods. In this table, the MAE value is calculated for the 6 objects in three different lighting conditions (indoor, sunny outdoor, and cloudy outdoor) that were considered for testing. The results show that the proposed method performs quite better compared to the previously developed physics-based methods (Smith et al., 2018; Mahmoud et al., 2012; Miyazaki et al., 2003). The MAE value on the whole dataset in these methods is between 41.44 and 49.03 degree, while the MAE value in the proposed method is equal to 18.06 degree on the whole dataset, which is approximately about half of the MAE value in the previous methods. Although the proposed method does not use any polarization information to reconstruct surface normal and only uses polarized images, the MAE value of this method is better than the DeepSfP (Ba et al., 2020) method and the MAE value for whole dataset in our proposed method is less than the DeepSfP method.

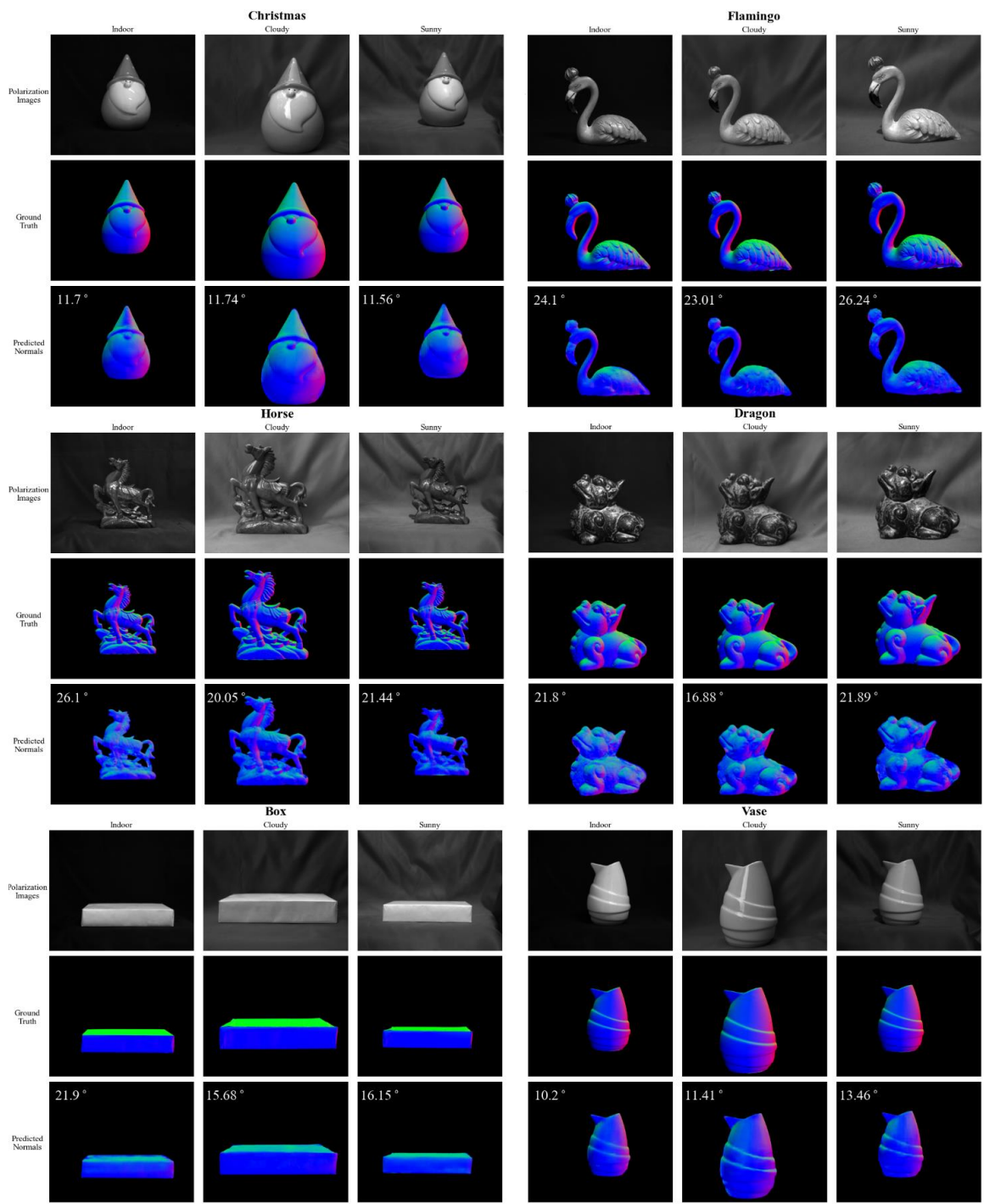

**Figure 6**. Reconstructed surface normal of objects in three different lighting conditions.

| Scene | Proposed | DeepSfP Ba et al., 2020 | Smith et al., 2018 | Mahmoud et al., 2012 | Miyazaki et al., 2003 |
|---|---|---|---|---|---|
| Box | **17.89°** | 23.31° | 31.00° | 41.51° | 45.47° |
| Dragon | **20.19°** | 21.55° | 49.16° | 70.72° | 57.72° |
| Father Christmas | **11.66°** | 13.50° | 39.68° | 39.20° | 41.50° |
| Flamingo | 24.43° | **20.19°** | 36.05° | 47.98° | 45.58° |
| Horse | 22.51° | **22.27°** | 55.87° | 50.55° | 51.34° |
| Vase | 11.69° | **10.32°** | 36.88° | 44.23° | 43.47° |
| Whole Set | **18.06°** | 18.52° | 41.44° | 49.03° | 47.51° |

Table 1. MAE values of objects across the three different lighting conditions. Other results are collected from the official results in DeepSfP (Ba et al., 2020).

## 5. CONCLUSION

In this paper we present a solution for estimating surface normal of objects using a convolutional neural network (U-Net). For geometric and semantic feature extraction, the ResNet18 has been utilized as the backbone of this network. Qualitative evaluation of the results showed that this method can reconstruct the surface normal in different lighting conditions including (indoor, sunny outdoor, and cloudy outdoor). Quantitative evaluation also showed that this method has the lowest MAE value compared to other physics-based methods, and also this method has significantly increased the accuracy of surface normal reconstruction.

The main benefit of this method is that it can estimate the surface normal of an object with just a series of polarized images taken by a polarization camera. This method is passive and does not have the active methods drawbacks. In addition, it also can reconstruct shiny objects and also objects in different lighting conditions such as indoor and outdoor lighting conditions. We plan to improve the network using polarization information in the future work.